\documentclass[acmsmall,screen,nonacm]{acmart}

\renewcommand\footnotetextcopyrightpermission[1]{} 
\AtBeginDocument{%
  \providecommand\BibTeX{{%
    \normalfont B\kern-0.5em{\scshape i\kern-0.25em b}\kern-0.8em\TeX}}}

\usepackage{xcolor}
\usepackage{tabularx}
\usepackage{enumitem}

\makeatletter
\newcommand{\enlabel}[2]{#2\def\@currentlabel{#2}\label{#1}}
\makeatother
\usepackage{amsthm}
\usepackage{acronym}
\usepackage{fbox}
\usepackage{hyperref}

\usepackage{float}
\floatstyle{plaintop}
\restylefloat{table}

\usepackage{tikz}

\usepackage{cleveref}

\theoremstyle{definition}
\newtheorem{example}{Example}

\sloppypar

\newcommand{\revA}[1]{#1} 
\newcommand{\revB}[1]{#1} 
\newcommand{\selfstar}{\mbox{self-*}}

\title{%
Artificial Collective Intelligence Engineering: a Survey of Concepts and Perspectives 
}

\author{Roberto Casadei}
\email{roby.casadei@unibo.it}
\orcid{0000-0001-9149-949X}
\affiliation{%
  \institution{{\sc{Alma Mater Studiorum--Universit{\`a} di Bologna}}}
  \streetaddress{Via dell'Universit{\`a} 50}
  \city{Cesena}
  \state{Italy}
}

\acrodef{ai}[AI]{artificial intelligence}
\acrodef{aci}[ACI]{artificial collective intelligence}
\acrodef{ac}[AC]{aggregate computing}
\acrodef{rq}[SQ]{\revB{scoping question}}
\acrodef{ci}[CI]{collective intelligence}
\acrodef{cas}[CAS]{collective adaptive system}
\acrodef{cps}[CPS]{cyber-physical system}
\acrodef{cci}[CCI]{computational collective intelligence}
\acrodef{ict}[ICT]{information and communication technology}
\acrodef{iot}[IoT]{Internet of Things}
\acrodef{ml}[ML]{machine learning}
\acrodef{marl}[MARL]{multi-agent reinforcement learning}
\acrodef{hmci}[HMCI]{human-machine collective intelligence}
\acrodef{hci}[HCI]{human-computer interaction}
\acrodef{lmas}[LMAS]{learning multi-agent system}
\acrodef{mas}[MAS]{multi-agent system}
\acrodef{sos}[SoS]{system-of-systems}
\acrodefplural{sos}[SOS]{systems-of-systems}
\acrodef{wsn}[WSN]{wireless sensor network}

\renewcommand{\cite}[1]{\citeauthor{#1} (\citeyear{#1})}

\begin{document}

\begin{abstract}
Collectiveness 
 is an important property
 of many systems---both natural and artificial.
By exploiting a large number of individuals,
 it is often possible to produce effects 
 that go far beyond the capabilities
 of the smartest individuals,
 or even to produce intelligent collective behaviour
 out of not-so-intelligent individuals.
Indeed, collective intelligence,
 namely the capability of a group to act collectively
 in a seemingly intelligent way,
 is increasingly often a design goal of engineered computational systems---motivated by
 recent techno-scientific trends like
 the Internet of Things, swarm robotics, and crowd computing, just to name a few.
For several years, 
 the collective intelligence 
 observed in natural and artificial systems
 has served as a source of inspiration
 for engineering ideas, models, and mechanisms.
\revA{Today,
 artificial and computational collective intelligence
 are recognised research topics, 
 spanning various techniques,
 kinds of target systems, 
 and application domains.
}
However, there is still a lot of fragmentation
 in the research panorama of the topic \revA{within computer science},
 and the verticality of most \revA{communities} and contributions
 makes it difficult to extract the core underlying ideas
 and frames of reference.
\revB{
The challenge is to identify, place in a common structure, and ultimately connect the different areas and methods addressing intelligent collectives.
}
To address this gap,
 this paper considers a set of \revB{broad scoping} questions
 providing a map 
 of collective intelligence research,
 mostly by 
 the point of view of computer scientists and engineers.
Accordingly, it covers preliminary notions, 
 fundamental concepts, 
 and the main research perspectives,
 identifying opportunities and challenges for 
 researchers on artificial and computational collective intelligence engineering.
\end{abstract}

\maketitle

\fancyfoot{}

\section{Introduction}
\label{sec:intro}

Nowadays, technical systems are evolving in complexity:
 they are increasingly large-scale, heterogeneous, and dynamic, posing several challenges to engineers and operators.
For instance, progress in the \ac{ict} 
 is promoting a future
 where computation is deeply integrated
 in a large variety of environments: our bodies,
 homes, buildings, cities, planet, and universe.
In other words,
 the vision of pervasive and ubiquitous computing
 is stronger than ever,
 with an increasing trend towards 
 the mass deployment of a large number of heterogeneous devices
 nearly everywhere,
 to improve existing applications and create new ones.
However, we still seem quite far at exploiting 
 the full potential 
 of the interconnected networks of devices at our disposal. 

Nevertheless, there is some progress. 
New paradigms and solutions have been proposed,
  often drawing from that powerful source of mechanisms and solutions that is nature.
Indeed, we are witnessing a long-term research endeavour
 aiming at bringing powerful properties and capabilities of living systems into technical systems~\citep{stein2021lifelike}.
Intelligence, evolution, emergence of novel capabilities, resilience, and social integration~\citep{stein2021lifelike,DBLP:journals/fgcs/BellmanBDEGLLNP21}
 are often observed in natural, living systems
 and considered important features of artificial, engineered systems as well.
Indeed, computer scientists and engineers
 are increasingly often 
 interested not just at making individual devices smarter,
 but also at making whole ecosystems of devices (and people) more collectively intelligent.
Creating \ac{ci} in artificial systems, however, is challenging.
\revA{
Indeed, various computer science and engineering fields 
 such as, e.g.,
 multi-agent systems~\citep{DBLP:books/daglib/0023784}
 and swarm robotics~\citep{DBLP:journals/swarm/BrambillaFBD13},
 have often encountered problems related to this ``\ac{ci} challenge''.
Moreover,
 the generality of the problem
 and the possibility of transferring ideas and techniques across fields
 has also motivated the emergence of a general research field specifically aimed at studying how to build \ac{ci} in artificial systems, also known as under terms \revB{such as} \emph{\ac{aci}}~\citep{DBLP:conf/aaai/ZhengYCZZWY18,wolpert1999introduction-ci-survey}
 and \emph{\ac{cci}}~\citep{DBLP:series/sci/2014-513}.
}

 
There exist some surveys on \ac{ci}/\ac{aci},
 \revB{but they tend to adopt specific viewpoints limiting the overall scope of the study,
 such as models for social computing systems~\citep{suran2020SLR-collective-intelligence}, 
 interaction modality~\citep{he2019ci-taxonomy-survey},
 or
 large-scale cooperative multi-agent systems~\citep{wolpert1999introduction-ci-survey}.
}
%
The main goal of this article 
 is to review the concepts, models, and perspectives
 needed for the \emph{engineering of \ac{aci}}.
We can say that the article mainly considers 
 \revA{\emph{cyber-physical collectives} as target systems,
 namely groups of interconnected computing devices, possibly situated in physical environment and possibly involving ``humans in the loop''~\citep{DBLP:journals/computer/SchirnerECP13}},
 which are to be thought 
 as ``programmable platforms'' for services and applications
 benefitting by the \ac{ci} emerging from their activity.
The idea is to provide a research map on \ac{ci} 
 for computer scientists and engineers\revA{, generally useful for the broad techno-scientific community}.
%
%
%

\revA{
In summary, we provide the following contributions:
\begin{itemize}
\item we perform a \emph{scoping review}~\citep{petticrew2008slr}, different from existing surveys \revB{in scope and focus}, covering concepts, models, and perspectives related to \ac{ci}, \ac{aci}, and their (software) engineering, which can also be seen as a foundation for more systematic reviews;
\item we provide a map and taxonomy of the \ac{aci} field, by connecting it with related fields and providing categories to frame research works on \ac{aci};
\item we outline opportunities and challenges for further research, in terms of target domains and interesting developments of existing methods.
\end{itemize}
\revB{
In other words, 
 we provide a broad overview of the field of \ac{ci}/\ac{aci},
 larger in scope and more oriented towards software systems engineering
 with respect to~\citep{malone2015handbook-collective-intelligence,suran2020SLR-collective-intelligence,he2019ci-taxonomy-survey,wolpert1999introduction-ci-survey}.
}
}

The article is organised as follows.
First, a set of broad \revB{scoping} questions are elicited,
 to provide a structure for the paper and its discussions.
After this, a survey of existing reviews relevant to \ac{ci} is presented, 
 to also motivate the perspective of this very article.
Then, the preliminary concepts of \revA{a ``collective'' and ``(individual) intelligence''} are briefly reviewed.
Upon this basis,
 to understand what \ac{ci} is,
 some reference definitions, examples, models, and classifications
 are reviewed from the literature.
Then,
 to discuss how \ac{ci} can be engineered,
 a number of perspectives are considered,
 under which some main approaches for \ac{ci} engineering are pointed out.
Building on such a presentation of approaches,
 a discussion of opportunities and challenges
 related to \ac{ci} engineering 
 is developed, providing directions for further research.
Finally, a wrap-up is provided with some conclusive thoughts.

\section{Method}
\label{sec:method}

\revB{
Our goal is to scope the large and fragmented area of (artificial) collective intelligence, in order to identify its key concepts, relevant perspectives, research problems, and gaps---with an emphasis on its engineering and its computational/\ac{ai} side.
Accordingly, we perform a scoping review~\citep{petticrew2008slr}.
This tool can be preferred over systematic reviews 
 whenever specific research questions are hard to identify or ineffective, or when the goal is to identify the \emph{types} of available evidence, clarify notions, and key characteristics/factors related to a concept~\citep{Munn2018}. 
Indeed, we seek to provide a map of the field, supporting more focussed and systematic reviews in the future.
}

We use a question-based method to drive the investigation and selection of the bibliography of this manuscript.
In particular, we consider the following \emph{\acp{rq}}.
\begin{enumerate}[label=\textbf{S.\arabic*}]
\item[\enlabel{rq-coll}{SQ0})] What is a collective?
\item[\enlabel{rq-intell}{SQ1})] What is (individual) intelligence?
\item[\enlabel{rq-ci}{SQ2})] What is collective intelligence?
\item[\enlabel{rq-examples}{SQ3})] What behaviours can be termed ``collective\revA{ly} intelligent''? Are there paradigmatic examples?
\item[\enlabel{rq-req}{SQ4})] What \revA{are the} requirements for collective intelligence?
\item[\enlabel{rq-indiv-vs-coll}{SQ5})] What relationship\revA{s exist} between individual and collective intelligence?
\item[\enlabel{rq-how}{SQ6})] How does collective intelligence unfold/emerge?
\item[\enlabel{rq-measure}{SQ7})] How \revA{can collective intelligence be measured}?
\item[\enlabel{rq-eng}{SQ8})] How \revA{can collective intelligence be built} artificially/computationally?
\item[\enlabel{rq-sota}{SQ9})] What is the state of the art of (computational) collective intelligence? 
\item[\enlabel{rq-field}{SQ10})] How is the research community on collective intelligence structured?
\end{enumerate}
\revB{
\ref{rq-coll}--\ref{rq-intell} cover the \emph{preliminary concepts} underlying the notion of \ac{ci}, setting the necessary background for addressing \ref{rq-ci}--\ref{rq-examples} (which are about \emph{what} \ac{ci} is)
 and \ref{rq-req}--\ref{rq-measure} (which are about the \emph{factors},  \emph{characteristics}, and \emph{mechanics} of \ac{ci}).
Then, \ref{rq-eng} is about the problem of \emph{engineering} \ac{ci},
and \ref{rq-sota}--\ref{rq-field} are meta-questions concerning research in the field.
%
%
Notice that these are broad scoping questions
 aimed mainly at providing directions for the search and identification of the research works included in the survey.
}

\section{Tertiary Study}\label{sec:tertiary-study}

In order to motivate the need for a survey on \ac{ci},
 we performed a tertiary study
 where secondary studies (e.g., surveys, systematic reviews) and collections are reviewed.
We organise these according to whether they consider \ac{ci} in its generality (i.e., abstracting from its applications and areas) 
 focus on its artificial/computational form (\acs{aci}/\acs{cci}),
 its swarm-like form,
 or specific kinds of collectives or goals.
Therefore, this section also provides a partial answer to \ref{rq-field}.

In general, we can observe a lack 
 of comprehensive reviews and maps of the \ac{ci} field.
From this situation, we draw a motivation for this article:
 providing a map of the topic, especially aimed at computer scientists and engineers, showing different perspectives and providing some highlights from the state of the art in \ac{aci}.

\subsubsection{Reviews on \ac{ci} as a general topic}
Two main surveys to date aim at addressing \ac{ci} as a general topic.
\cite{he2019ci-taxonomy-survey}
 analyse \ac{ci} across different fields 
 based on a taxonomy that distinguishes between isolation, collaboration, and feedback-based \ac{ci} paradigms.
%
\cite{suran2020SLR-collective-intelligence}
 performed a systematic literature review 
 to elicit a general model of \ac{ci} and its attributes\revB{, with a focus on social computing}.
These two contributions consider, integrate, and somewhat subsume previous, more limited, or less general \ac{ci} models and reviews \citep{yu2018review-ci-crowd-science-perspective,DBLP:conf/medes/LykourentzouVL09,DBLP:journals/corr/abs-1204-3401,krause2010animals-humans}, 
and so they will be discussed more extensively in later sections.
However, \revB{to the best of our knowledge, there are no comprehensive mapping studies providing a broad overview of the field for computer scientists}.

\subsubsection{Multi-disciplinary collections}

In \citep{malone2015handbook-collective-intelligence}, 
 essays on \ac{ci} are collected from different fields 
 including
 economics, biology, \ac{hci}, \ac{ai}, organisational behaviour, and sociology.
In~\citep{DBLP:journals/corr/abs-2112-06864},
 a collection of contributions 
 from a workshop gathering scientists in different areas
 is provided,
 with the goal of sharing \emph{``insights about how intelligence can emerge from interactions among multiple agents---whether those agents be machines, animals, or human beings.''}

\subsubsection{Reviews on \ac{aci}/\ac{cci}}
The paper by \cite{wolpert1999introduction-ci-survey},
 published in 1999,
 surveys \ac{cci} systems
 across the categories of 
 (i) \ac{ai} and \ac{ml}, including \acp{mas};
 (ii) social science-inspired systems, such as those found in economics and game theory;
 (iii) evolutionary game-theoretical approaches;
 (iv) biologically-inspired systems, like swarm intelligence, artificial life, and population approaches;
 (v) physics-based systems;
 and
 (vi) other research fields ranging from network theory and self-organisation.
This is a very rich survey but covers research published before the year 2000 \revA{and is slightly focused towards automatic and utility-based approaches}.

The editorial by \cite{DBLP:journals/fgcs/Jung17}
 reviews special issue papers on the integration of \ac{cci} and big data, 
 where it is considered how data-driven \ac{ci} can help in (i) collecting data, (ii) analysing data, and (iii) using data e.g. to support decision making.

The review by \cite{DBLP:journals/corr/abs-1803-05464}
 provides a survey and taxonomy
 of multi-agent algorithms for collective behaviour,
 classified into: consensus, artificial potential functions, distributed feedback control, geometric algorithms, state machines and behaviour composition, bio-inspired algorithms, density-based control, and optimisation algorithms.
What emerges is a rather sharp distinction between low-level (e.g., bio-inspired self-organisation) 
 and high-level coordination.

\subsubsection{Reviews on swarm intelligence}

Several reviews on swarm intelligence have been published~\citep{Chakraborty2017swarm-review-algorithms,DBLP:journals/swevo/MavrovouniotisL17,DBLP:journals/swevo/RajasekharLDS17,DBLP:journals/ijbic/YangH13,DBLP:journals/swevo/FisterFYB13,dorigo2019aco-overview-recent-advances,DBLP:journals/swevo/NguyenXZ20,DBLP:journals/eaai/FigueiredoMSSGF19,DBLP:journals/eaai/ZhangLCCW15,navarro-2013-survey-collective-movement-robots,DBLP:journals/jpdc/ZedadraGJSSF18,DBLP:journals/thms/KollingWCS016,DBLP:journals/swarm/BrambillaFBD13,DBLP:journals/swevo/SchranzCSEASS21}.
In the swarm intelligence field,
 a large part of research is devoted to devising
 (meta-)heuristics and algorithms 
 for solving complex optimisation problems.
\cite{DBLP:journals/swevo/MavrovouniotisL17} focus on swarm algorithms for dynamic optimisation, namely in settings where the environment changes over time.

Moreover, reviews in this context
 often adopt an angle based on 
 what natural system inspired swarm intelligence mechanisms.
For instance,
 \cite{DBLP:journals/swevo/RajasekharLDS17}
 provide a survey on algorithms
 inspired by honey bees,
 e.g. based on mating, foraging, and swarming behaviours
 of honey bees;
 similar surveys exist for bat algorithms~\citep{DBLP:journals/ijbic/YangH13}, 
 firefly algorithms~\citep{DBLP:journals/swevo/FisterFYB13},
 ant colony optimisation~\citep{dorigo2019aco-overview-recent-advances}.

\revA{Some surveys consider swarm intelligence applied to specific problems such as
self-organising pattern formation~\citep{DBLP:journals/ras/OhSSJ17}, }
feature selection~\citep{DBLP:journals/swevo/NguyenXZ20}, 
clustering~\citep{DBLP:journals/eaai/FigueiredoMSSGF19},
green logistics~\citep{DBLP:journals/eaai/ZhangLCCW15},
collective movement~\citep{navarro-2013-survey-collective-movement-robots}.
\revA{Other surveys consider swarm intelligence in particular contexts or as exhibited by particular kinds of systems, such as
\ac{iot} systems}~\citep{DBLP:journals/jpdc/ZedadraGJSSF18},
\revA{\acp{cps}~\citep{DBLP:journals/swevo/SchranzCSEASS21},}
and robot swarms~\citep{DBLP:journals/swarm/BrambillaFBD13,DBLP:journals/thms/KollingWCS016}.

\subsubsection{Reviews on \ac{ci} for specific systems and settings}
\label{ssec:reviews-ci-specific}

Reviews from specific viewpoints
 include collections and surveys 
 on human \ac{ci}~\citep{DBLP:journals/corr/abs-1204-3401},
 deep learning~\citep{DBLP:journals/corr/abs-2111-14377},
 enterprise information systems~\citep{DBLP:journals/eis/NguyenHS19},
 \revA{ and
 sociotechnical systems supported by 5G communications~\citep{DBLP:journals/access/NarayananKMHSSG22}%
}.

\cite{DBLP:journals/corr/abs-1204-3401}
 performed a literature review of \ac{ci} in human context,
 grouping contributions into 
 (i) micro level, emphasising enabling factors;
 (ii) emergence (or meso) level, emphasising how global patterns arise from local activity; and
 (iii) macro level, emphasising the kinds of system output.
A review on human \ac{ci}
 by a crowd science perspective 
 is provided by \cite{yu2018review-ci-crowd-science-perspective}.
\cite{krause2010animals-humans}
 review and compare swarm intelligence in animals and humans.

\cite{DBLP:journals/corr/abs-2111-14377}
 performed a survey of recent developments
 on the embedding of \ac{ci} principles into deep learning methods.
They discuss e.g. how \ac{ci} can help in devising novel architectures and training algorithms, and recent works  on multi-agent (reinforcement) learning.
Studies like this one are important since they elicit 
 and strengthen trans-disciplinary relationships
 which are key for complex interdisciplinary fields like \ac{ci}.
 
\revA{
\cite{DBLP:journals/access/NarayananKMHSSG22} provide a survey of the \ac{ci} emerging in human-machine socio-technical systems supported by 5G communications.
The discussed applications include 
 road traffic control, unmanned aerial vehicles, smart grid management, and augmented democracy.
The point is that to realise their full potential,
 these kinds of decentralised socio-technical systems
 often require proper connectivity properties and capabilities
 to support and foster the emergence of \ac{ci}.
For instance, from the analysis, the authors foresee that the 5G communication technology can promote \ac{ci} by enhancing aspects like connectivity with neighbour nodes, interaction protocols, knowledge exchange, and the exploration-exploitation tradeoff via improved speed, latency, and reliability.
On the other hand, there are significant challenges addressed by current and hopefully by future research in terms of security, privacy, and radio resource management.
}

\section{Preliminary Concepts}
\label{sec:preliminary}

This section provides an introduction to \revA{the notions of} collectives and individual intelligence, hence addressing \ref{rq-coll} and \ref{rq-intell}\revA{,
and providing preliminary concepts for introducing and discussing \ac{ci} in the next section}.

\subsection{Collectives}\label{s:collectives}

Informally, a \emph{collective} is a (possibly dynamic) group of largely \emph{homogeneous} individuals, which are also called the \emph{members} of the collective.
Different works may use different or more specific definitions for a collective.
Different fields often target different kinds of collectives, often resulting in implicit assumptions.

Devising a general and comprehensive characterisation of collectives 
 is an open research problem,
 addressed in the context of 
 \emph{mereology}, namely the study of \emph{parthood} relations, 
 and \emph{ontology}, namely the study of ``what there is''.
In the literature, a few formal theories attempt to deeply characterise collectives and collective phenomena~\citep{brodaric2020pluralities-collectives-composites,galton2016extensional-intensional-collectives,wood2009taxonomy-collective-phenomena,bottazzi2006intentional-collectives}.

For instance, in \citep{wood2009taxonomy-collective-phenomena},
 a taxonomy of collective phenomena is provided,
 along the classification criteria
 of \emph{membership} (concerned with the identity and cardinality of the members of a collective),
 \emph{location} (of the collective as well as of its members),
 \emph{coherence} (the source of ``collectiveness''),
 \emph{roles} (if members are distinguished by roles),
 \emph{depth} (concerning levels of collectives).
In particular, two main sources for collectiveness
 can be devised:
 internal or external \emph{causes},
 and \emph{shared purposes} or \emph{goals}.
Regarding depth, it is worth noticing that, unlike the \emph{componenthood} relation in composites, \emph{membership} in collectives is generally not transitive~\citep{brodaric2020pluralities-collectives-composites}.
Composites can be defined as structured pluralities or groups of parts, called \emph{components}, playing specific functions~\citep{brodaric2020pluralities-collectives-composites}.
In the literature,
 it is generally assumed that 
 composites are heterogeneous,
 \revA{while} collectives are homogeneous~\citep{brodaric2020pluralities-collectives-composites}.
 
\revA{
Moreover, a collective is often intended to be a ``concrete particular'' (i.e., not an abstraction like a mathematical set)
and a ``continuant'' (i.e., a particular existing and possibly changing over a time span)~\citep{wood2009taxonomy-collective-phenomena}.
Defining a general, comprehensive, and precise characterisation or taxonomy of collectives is not trivial~\citep{wood2009taxonomy-collective-phenomena}.
For instance, certain collectives may require a certain number of members or roles to be filled to exist~\citep{wood2016considering-collectives}, or may change identity following certain changes in their composition.
Sometimes, collectives may be abstracted 
 by specific collective properties
 or collective knowledge~\citep{DBLP:journals/cas/Nguyen08}.
Collectiveness may also be considered as a degree, 
 and hence a quantifiable property~\citep{Daniels2016}
 of phenomena and groups of individuals.
}

\begin{table}
\begin{tabularx}{\columnwidth}{|p{0.25\columnwidth}|p{0.25\columnwidth}|X|}
\hline
\textbf{Concept} & \textbf{(Typical) Parent concept} & \textbf{(Typical) Defining properties}
\\\hline
Plurality; Collection; Group; Set & & Set-inclusion
\\\hline
Composite & Plurality & Componenthood, heterogeneity
\\\hline
Collective & Plurality & Membership, homogeneity
\\\hline
Crowd & Collective & Nature (humans)
\\\hline
Swarm & Collective & Nature (insect-like)
\\\hline
Robot swarm & Swarm & Nature (simple robots), structure (high numbers)
\\\hline
Herd; Flock; School & Collective & Nature (animals)
\\\hline
Organisation & Composite / Collective & Structure, roles
\\\hline
System & Composite / Collective & Interacting elements; Boundary
\\\hline
Multi-agent system & System & Nature (agency)
\\\hline
\end{tabularx}
\caption{Common group-like notions addressed in computer science and engineering.}
\label{table:group-concepts}
\end{table}

There exist several related group-like notions,
 which differ e.g. by perspective, the key relation between items, or the fundamental property of the group. 
Some of these group-like notions are summarised in \Cref{table:group-concepts}\revA{, with a proposed classification---following~\cite{brodaric2020pluralities-collectives-composites}}, though different meronomies are possible.
A collective is a particular kind of plurality or group. 
%
%
Crowds, swarms, herds, flocks, schools can generally be considered specific kinds of collectives.
\revA{
Organisations and systems might be modelled as constructs based on the structural arrangement and heterogeneity of composites,
 but are also amenable to be characterised as collectives.
}

Like for the notion of intentional stance~\citep{dennett1989intentional-stance},
 it may make sense to adopt a \emph{collective stance}
 in which e.g. \emph{``the human species [a group] is viewed as a single organism''}~\citep{gaines1994collective-stance}\revA{, though the idea of \emph{collective intentionality} is problematic and subject of intense philosophical debate~\citep{Schweikard2021encyclopedia-collective-intentionality}}.
Indeed, we believe that the perspective of collectiveness
 can provide a complementary point of view to that of an individual
 for understanding and engineering various sorts of systems involving groups of individuals.
However, when addressing themes involving collectives (such as \ac{ci}),
 it is important to clarify what kind of collectives are addressed, 
 as this would help to clarify the assumptions and generality of a specific contribution.
%

%

\subsection{Intelligence}\label{s:intelligence}

Intelligence is a controversial and elusive concept subject to philosophical debate~\citep{legg2007collection-defs-intelligence},
 best understood as a nomological network of constructs~\citep{reeve2011nature-and-structure-of-intelligence}.
Etymologically, intelligence comes from Latin ``intelligere'', which means ``to understand''.
It can be defined as \emph{``the global capacity of the individual to act purposefully, to think rationally, and to deal effectively with the environment''}~\citep{wechsler1944measurement-adult-intelligence},
or the property that \emph{``measures an agent's ability to achieve goals in a wide range
of environments''}~\citep{legg2007collection-defs-intelligence}.
In general, there are two different \revB{interpretations}:
 intelligence as either
 a collection of task-specific skills or a general learning ability~\citep{chollet2019measure-intelligence},
 which reflect the distinction between \emph{crystallised} and \emph{fluid} abilities, respectively.

Problems about intelligence
 include, for instance, 
 its
 definition and modelling, such as devising the structure of intelligence~\citep{reeve2011nature-and-structure-of-intelligence};
 its relation with action;
 its measurement and evaluation;
 its analysis;
 and its construction and development.

Concerning the theories of intelligence,
 there are two main traditions~\citep{reeve2011nature-and-structure-of-intelligence}:
 the \emph{psychometric tradition}, 
 based on the number and nature of basic cognitive abilities or \emph{factors};
 and the developmental or holistic perspective,
 based on acquired intellect.

The problem of the \emph{measure} of intelligence~\citep{hernandez2017evaluation-in-ai,chollet2019measure-intelligence}
 is of course related to what representation or model of intelligence is considered,
 and is complicated 
 by the need of distinguishing between causality and correlation,
 selecting a representative set of environments for evaluation, etc.
Carrol defines an \emph{ability} (i.e. an intelligence factor) as a source of variance in performance for a certain class of tasks~\citep{carroll1993human-cognitive-abilities}.
Measuring intelligence is based on \emph{factor analysis}, i.e., it works by running specific tests (\emph{observables})
and using factors (\emph{unobservables}) as possible explanations for correlations among the observables, describing their variability. 
It is expected that 
 the nature of the entity
 whose intelligence we are considering
 would drive and require the definition
 of suitable factor models.

\revA{
Various taxonomies of intelligence have been proposed over time.
A common distinction is between \emph{natural}~\citep{DBLP:journals/ficn/Gerven17} and \emph{artificial intelligence}~\citep{DBLP:books/aw/RN2020}.
Both can be considered under the unifying notion of \emph{abstract intelligence}~\citep{DBLP:journals/ijssci/Wang09}.
}

%

\section{Understanding Collective Intelligence}
\label{sec:ci-what}

\revA{On the basis of the preliminary concepts introduced in the previous section, t}his section focusses on \emph{what} \ac{ci} is, according to literature, \revA{discussing definitions, examples, models, and the main classifications of \ac{ci} (namely \ac{aci} and \ac{cci}) we are interested into,} 
hence addressing \ref{rq-ci} to \ref{rq-measure}. 
\revA{
Understanding the goals, characteristics, and main frames of reference of \ac{ci} is important before turning to the problem of \ac{cci} engineering in the next section.
}

\subsection{Definitions \revA{and characterisations of collective intelligence}}\label{ssec:ci-defs}

\Acl{ci} is the intelligence that can be ascribed to a collective---where a collective is a multiplicity of entities \revA{(commonly characterised as discussed in the previous section)}.
Indeed, by abstracting a collective as a \emph{whole}, namely as \revA{a \emph{higher-order individual} in turn (consisting of other individuals, which are its \emph{members})},
it should be possible to transfer
 characterisations of individual intelligence to it.
%

\Cref{table:ci-defs} reports some definitions of \ac{ci} taken from the literature.
From them, it is possible to see 
 recurrent as well as peculiar aspects of \ac{ci} characterisations.

\subsubsection{Reuse of (individual) intelligence \revA{definitions}.} Some definitions do not attempt to re-define ``intelligence'' 
 but merely bring existing characterisations of intelligence, commonsense acceptations, or its general meaning as a nomological network of concepts~\citep{reeve2011nature-and-structure-of-intelligence} to the collective realm.
 This has the advantages of simplicity, generality, and \emph{openness}, 
  which may promote multi-, inter- and trans-disciplinarity.

\subsubsection{General vs. task-specific.} If we reuse existing notions of intelligence, it means that we may consider how different definitions in turn apply to collective entities. For instance, similarly to individual intelligence, \ac{ci} may be considered as a general problem-solving ability or as a set of specific skills.
\revA{
Evidence for the existence of a general \ac{ci} statistical factor $c$ in human groups has been provided by \cite{Woolley2010evidence-ci-factor},
 where such factor is shown to be more correlated with average social sensitivity and diversity, rather than with average or maximum individual intelligence of the members.
}

\subsubsection{Collectives of different natures.}
 Some definitions largely abstract from the nature of collectives (cf. ``collections'' or ``groups of individuals'', ``artificial and/or natural''), 
 some assume a minimal set of characteristics for individuals (cf. agency, ability to interact, etc.),
 some require that the individuals are connected in some way (cf. interaction, or existence of social structures).
 
\subsubsection{Different sources for collectiveness and mechanisms for \ac{ci}.}
 Terms like interaction, collaboration, competition, and social structure
 might be used to further constrain the scope of \ac{ci}
 to particular kinds of collectives
 or to different mechanisms thereof that are possible
 for supporting \ac{ci}.

\subsubsection{Connection to emergence.} Various definitions build on the notion of \emph{emergence}, which 
 relates to the production, in a system, of radically novel, coherent macro-level patterns from micro-level activity~\citep{DBLP:conf/atal/WolfH04}.
 
\subsubsection{Phenomenological approach.} \revA{Similarly to emergence, which is often studied phenomenologically~\citep{Minati2018,engineering-emergence2018}}, 
some \ac{ci} definitions 
 adopt a phenomological standpoint where the focus is not on what \ac{ci} actually is, but on the phenomena that may be associated to it.

\revA{
\subsubsection{Positive vs. negative \ac{ci}.}
It is common to consider \ac{ci} as a \emph{quantifiable} property and specifically as a \emph{signed} quantity, i.e., positive or negative. Indeed, various authors talk about \emph{negative} collective intelligence~\citep{szuba2001computational-collective-intelligence,DBLP:journals/firai/LaanMP17} in order to characterise the cases where a collective would perform worse than one of its individual members. In such cases, the social constraints effectively hinder individual abilities with no benefit. 
}

%

{
\begin{table*}
\newcommand{\bo}[1]{#1}
\begin{tabularx}{\textwidth}{|p{0.14\textwidth}|X|p{0.3\textwidth}|}
\hline
\textbf{Ref.} & \textbf{Definition} & \textbf{Remarks}
\\\hline
\cite{malone2015handbook-collective-intelligence} & \emph{Groups of individuals \bo{acting collectively} in ways that seem intelligent} &
\begin{itemize}[nosep,leftmargin=*]
\item ``Reuse'' of the notion of intelligence
\item Collective action
\end{itemize}
\\\hline
\cite{DBLP:conf/iccci/2009} & \emph{The form of intelligence that \bo{emerges} from the \bo{collaboration} and \bo{competition} of many individuals (artificial and/or natural)} & 
\begin{itemize}[nosep,leftmargin=*]
\item Emergence
\item Mechanisms (collaboration, competition)
\item Members of different nature
\end{itemize}
\\\hline
\cite{he2019ci-taxonomy-survey} & 
\emph{Collective intelligence (CI) refers to the intelligence that \bo{emerges} at the \bo{macro-level} of a
collection and transcends that of the individuals.}
&
\begin{itemize}[nosep,leftmargin=*]
\item Emergence (transcendence)
\item Levels (macro, micro)
\end{itemize}
\\\hline
\cite{wolpert1999introduction-ci-survey} & \emph{[COllective INtelligence (COIN)] Any pair of a large,
distributed collection of \bo{interacting computational processes} among which there is little
to no centralized communication or control, together with a ``world utility'' function
that rates the possible dynamic histories of the collection.} &
\begin{itemize}[nosep,leftmargin=*]
\item Requirements (interaction, decentralisation)
\item Embedded metric
\end{itemize}
\\\hline
\cite{szuba2001computational-collective-intelligence} & \emph{We can say that the phenomenon of CI has emerged in a \bo{social structure} of interacting \bo{agents} or \bo{beings}, over a certain period, iff the \bo{weighted sum of problems} they can solve together as a social structure is higher during the whole period than the sum of problems weighted in the same way that can be solved by the agents or beings when not interacting}
&
\begin{itemize}[nosep,leftmargin=*]
\item Requirements (social structure, interaction)
\item Embedded metric
\item Dynamic property
\item Negative and positive \ac{ci}
\end{itemize}
\\\hline
\cite{DBLP:conf/medes/LykourentzouVL09} & \emph{Collective intelligence (CI) is an emerging research field which aims at combining human
and machine intelligence, to improve community processes usually performed by large
groups.} & \begin{itemize}[nosep,leftmargin=*]
\item Hybrid or human-machine \ac{ci} 
\end{itemize}
\\\hline
\end{tabularx}
\caption{Some definitions of \ac{ci} from the literature.}\label{table:ci-defs}
\end{table*}
}

\subsection{Examples}

In the following, notable examples of \ac{ci} are briefly reviewed.

\begin{example}[Markets]
Markets are economic systems that consist of a large number of rational self-interested agents, buyers and sellers, 
 that engage in transactions regarding assets.
The prices of assets change to reflect supply and demand, as well as the larger context, and can be seen as a reification of the collective intelligence of the entire market~\citep{lo2015markets}.
So, markets can be seen as a mechanism for sharing information and making decisions about how to
allocate resources in a collectively intelligent way~\citep{malone2015handbook-collective-intelligence}.
Accordingly, market-based abstractions have been considered in computer science to promote globally efficient systems~\citep{DBLP:conf/sigopsE/MainlandKLPW04}.
\end{example}

\begin{example}[Wisdom of crowds]
Crowds -- groups of people -- can be of different kinds (cf. physical vs. psychological crowds)
 and can exhibit different degrees of \ac{ci}.
A crowd can \revA{exhibit intelligent~\citep{surowiecki2005wisdom-crowds} or unintelligent behavior~\citep{DBLP:journals/firai/LaanMP17}}.
\cite{surowiecki2005wisdom-crowds}
 popularised
 the term ``wisdom of crowds'',
 showing that groups are able of good performance
 under certain circumstances, 
 providing aggregate responses that incorporate and exploit the collective knowledge of the participants.
Among the conditions required for a crowd to be wise,
 \cite{surowiecki2005wisdom-crowds}
 identified
 \emph{diversity} (of individuals),
 \emph{independence} (of individual opinions),
 and 
 \emph{decentralisation} (of individual knowledge acquisition)\revA{---whose importance has been confirmed by later studies such as, e.g., those by \cite{Woolley2010evidence-ci-factor}}.
\end{example}

\begin{example}[Swarm intelligence]
Swarm intelligence is the \ac{ci} that emerges 
 in groups of simple agents~\citep{bonabeau1999swarmintelligence-book}.
Swarm intelligence was first observed in natural systems, such as insect societies (e.g., ant colonies, beehives),
 which inspired mechanisms and strategies
 for improving the flexibility, robustness, and efficiency
 of artificial systems.
With respect to the general field of \ac{ci},
 swarm intelligence may be considered as a sub-field
 that deals with very large groups
 and individuals behaving according to simple rules.
Since the criteria of cardinality and simplicity 
 are degrees, the boundaries of the field is also fuzzy.
\end{example}

\begin{example}[\Aclp{lmas}]
Another notable example of \ac{ci} 
 is given by \acp{mas}~\citep{DBLP:books/daglib/0023784}.
Unlike swarms, \acp{mas}
 usually comprise rational agents,
 possibly structured into organisations,
 and possibly exhibiting properties of strong agency~\citep{DBLP:books/daglib/0023784}, i.e., 
 which may in turn be individually intelligent.
The agents as well as the \ac{mas} 
 may be able to \emph{learn}
 about the environment,
 themselves,
 or the behaviour that they should follow
 to maximise some local or global notion of utility~\citep{wolpert1999introduction-ci-survey}.
\end{example}

\begin{example}[\Acl{hmci}]
A powerful example of \ac{ci}
 is the so-called \emph{\acf{hmci}}~\citep{smirnov2019human-machine-ci}
 or \emph{hybrid \ac{ci}}~\citep{DBLP:journals/ais/PeetersDBBNSR21,moradi2019collective-hybrid-intelligence},
 which is the one that applies
 to heterogeneous systems involving 
 both machines and humans.
The idea is to promote the synergy 
 between artificial/machine intelligence
 and human intelligence,
 which are often seen as complementary forms of intelligence.
An exemplar of \ac{hmci} is Wikipedia, 
 a hypermedia system 
 of interconnected collective knowledge,
 which is created and revised by humans 
 through the mediation of Web technologies.
Wikipedia data can also be autonomously processed by agents
 to build other kinds of applications leveraging 
 its collective knowledge.
\end{example}

%

\subsection{Models}

Here, we briefly review two main general models of \ac{ci} from the literature, which comprehensively summarise and integrate previous models.

\subsubsection{The isolation-, collaboration-, and feedback-based \ac{ci} paradigms~\citep{he2019ci-taxonomy-survey}}
\cite{he2019ci-taxonomy-survey} propose a taxonomy of \ac{ci} into three paradigms
  of increasing power,
  based on the absence or presence 
  of \emph{interaction} and \emph{feedback} mechanisms.
In their view, \ac{ci} can be generally regarded as an aggregation of individual behaviour results.
Then:
\begin{enumerate}
\item \emph{Isolation paradigm.} The individuals are isolated and behave independently, producing results that are aggregated in some way. The aggregation result does not affect the individual behaviours. Isolation studies use statistical and mining tools.
\item \emph{Collaboration paradigm.} There is direct or indirect \emph{interaction} between the individuals. Indirect interaction can be modelled through a notion of \emph{environment}. Aggregation operates on individual behaviour results and the environment state. The aggregation result does affect neither the individual behaviours nor the environment.
\item \emph{Feedback paradigm.} This paradigm adds to the interaction paradigm a ``downward causation'' of the aggregation result on the individual behaviours and/or the environment.
\end{enumerate}

\subsubsection{\ac{ci} framework by \cite{suran2020SLR-collective-intelligence}}
\cite{suran2020SLR-collective-intelligence} analyse 12 studies on \ac{ci} 
 and devise a \emph{generic} model
 based on 24 \ac{ci} attributes
 split into 3 \ac{ci} components: individuals, coordination/collaboration activities, and communication means.
The generic model
 is based on:
\begin{itemize}
\item Characterisation of \emph{who} is involved in a \ac{ci} system, in terms of: passive actors (users);
 active actors (\ac{ci} contributors), which may be crowds or hierarchies;
 properties of actors in terms of diversity, independence, and critical mass;
 and 
 interactions.
\item Characterisation of \emph{motivation} of \ac{ci} actors: intrinsic or extrinsic.
\item Characterisation of \ac{ci} goals: individual and community objectives.
\item Characterisation of \ac{ci} processes: in terms of types of activities (decide, contest, and voluntary) and interactions (dependent or independent).
\end{itemize}
Moreover, \ac{ci} systems
 can be considered as complex adaptive systems
 and
 often are subject to requirements
 for proper functioning
 e.g. on state, data, aggregation, decentralisation, task allocation, and robustness.


%
%
%
%
%
%
%

\revA{
\subsection{Factors and quantification of \ac{ci}}

Key scientific questions,
fundamental for both understanding and engineering \ac{ci},
 include
 what \emph{factors} promote or inhibit \ac{ci},
 and,
 specifically,
 what is the relationship between individual and collective intelligence.
We already mentioned the seminal work by \cite{Woolley2010evidence-ci-factor}
 sustaining the idea of a general \ac{ci} factor $c$,
 shown to be more correlated with the level of sociality than with the levels of intelligence of individuals.
We also pointed out the example of swarm intelligence 
 as a kind of \ac{ci} emerging from a multitude of 
 simple agents characterised by limited individual intelligence.
In this example, clearly, it is the aspect of \emph{interaction} -- with other agents and/or the environment -- that fosters the production of effective patterns of behaviour.

Works have been carried out to investigate these relationships.
For instance, in a later study,
\cite{woolley2015collective-intelligence-and-group-performance} focus on (i) group composition, e.g., in terms of skills and diversity of the members of a group;
and (ii) group interaction, e.g., in terms of structures and norms constraining and ruling the interaction.
They found that the individual skills that contribute the most to \ac{ci} are those that bring sufficient diversity and effectiveness in collaboration,
 whereas group-level psychological elements like satisfaction and cohesiveness are not influential.
Considering different kinds of interactive cognitive systems, 
\cite{DBLP:conf/ecai/ChmaitDLGI16} study the influence of the following factors: 
(i) concerning individuals:
 individual intelligence,
 individual reasoning/learning speed;
(ii) concerning cooperation: 
 cardinality of the collective,
 time to interact,
 communication protocol; and
(iii) concerning agent-environment interaction: 
 search space complexity (through uncertainty),
 and algorithmic complexity of the environment.
They quantify the \ac{ci} of a group of agents
 as the mean accumulated reward in a set of test environments, hence extending the Anytime Universal Intelligence Test~\citep{HernndezOrallo2010} to collectives.
What is observed is that such factors -- considered independently and/or in joint configurations with other factors -- do shape the \ac{ci} of groups in non-trivial ways.
These factors are also related to the components of \ac{ci} models---a nice overview is provided in~\citep{suran2020SLR-collective-intelligence}.
}

\subsection{Main kinds of \acs{ci}}
\label{ssec:main-kinds}

A typical classification of \ac{ci}
 is by the nature of the entities involved.

\subsubsection{Natural \ac{ci}}
Natural \ac{ci} 
 is the \ac{ci} exhibited by 
 collectives found in nature,
 such as swarms of insects,
 packs, herds, or groups of animals,
 crowds of people,
 flocks of birds,
 schools of fishes, etc. 
In all these systems
 there exist 
 non-trivial
 collective phenomena
 and societal aspects 
 that deserve deep investigation.
Insect societies are analysed e.g. in the seminal book by ~\citep{bonabeau1999swarmintelligence-book}.
For collective animal behaviour, 
 one of the main references is the book 
 by \cite{sumpter2010collective-animal-behavior},
 which describe collective phenomena
 as those in which \emph{``repeated interactions among many individuals produce patterns on a scale larger than themselves''}.
For \ac{ci} in humans, 
 some historical references include~\citep{lebon2002crowd,surowiecki2005wisdom-crowds};
 moreover, there are contributions
 for specific human settings like 
 crowds of pedestrians~\citep{10.1371/journal.pone.0177328}
 and problems like
 e.g. the relationship between language and collective action~\citep{smith2010communication-collective-action}. 

The study of natural \ac{ci}
 is important because 
 it is powerful source of inspiration
 for \ac{ci} mechanisms
 to be applied to artificial systems~\citep{bonabeau1999swarmintelligence-book}.
 
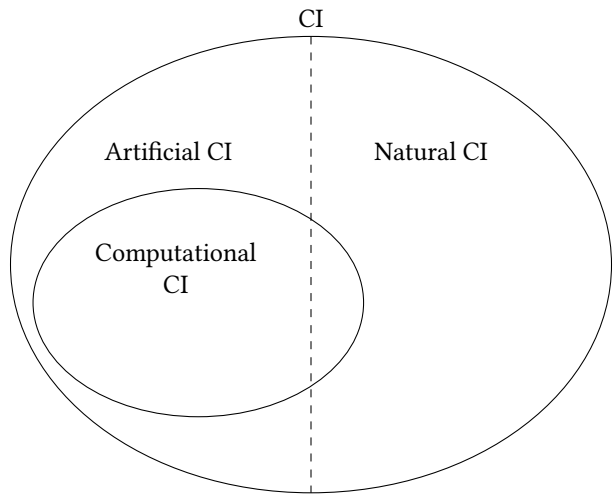
\begin{figure}
  \def\circleci{(2.0, 2.0) circle [x radius=4cm, y radius=3cm]}
  \def\circleaci{(0.8, 2.0) circle [x radius=2.3cm, y radius=2.0cm]}
  \def\circlenci{(3.2, 2.0) circle [x radius=2.3cm, y radius=2.0cm]}
  \def\circlecci{(0.5, 1.5) circle [x radius=2.2cm, y radius=1.5cm]}
    \begin{tikzpicture}
      \draw \circleci; 
      \draw (2.0,5.0) node [text=black,above] {CI};
      \draw (0.1,3.25) node [text=black,above] {Artificial CI};
      \draw (3.6,3.25) node [text=black,above] {Natural CI};
      \draw[dashed] (2.0,5.0) -- (2.0,-1.0);
      \draw[] \circlecci; 
      \draw (0.2,1.5) node [text=black,above,align=center] {Computational\\CI};
    \end{tikzpicture}
\caption{\revA{The relationships between \acf{ci}, \acf{aci}, and \acf{cci}. The dashed line is used to denote the false dichotomy between artificial and natural \ac{ci}.}}
\label{fig:aci-vs-cci}
\end{figure}

\subsubsection{Artificial (\acs{aci}) and Computational Collective Intelligence (\acs{cci})} 
\Ac{aci} is the \ac{ci} exhibited by \revA{human-made} machines.
\revA{Notice that, strictly speaking, natural and artificial \ac{ci} constitute a false dichotomy since there is inherent subjectivity regarding where the line between the two is drawn, and these could also be considered as a gradation.}
\revA{\ac{aci} and \ac{cci} are mostly considered as synonyms in the literature.
However, some authors refer to \ac{cci} as a particular kind of \ac{aci}, i.e.,} \emph{``as an AI sub-field dealing with soft computing methods which enable making
group decisions or processing knowledge among autonomous units acting in distributed environments''}~\citep{DBLP:series/sci/2014-513}.
\revA{
\emph{Soft computing} methods are those that help to address complex problems by overcoming approximation and uncertainty,
 using techniques such as fuzzy logic, expert systems, machine learning, genetic algorithms, artificial neural networks~\citep{Ibrahim2016soft-computing}.
In other words, the distinction between \ac{aci} and \ac{cci} may follow the common way to distinguish between artificial and computational intelligence~\citep{engelbrecht2007computational-intelligence-intro}, where the former tends to prefer hard, symbolic approaches while the latter tends to prefer soft and bio-inspired computing techniques.
}
\revA{The relationships between \ac{ci}, \ac{aci}, and \ac{cci} is shown in \Cref{fig:aci-vs-cci}.
\ac{cci} might also be intended as a part of natural \ac{ci} 
 to account for the notion of \emph{biological computation}~\citep{DBLP:journals/ubiquity/Mitchell11},
 whereby biological systems are considered as computing devices~\citep{DBLP:journals/ficn/Gerven17}.
However, not all of \ac{aci} is necessarily computational, since also mechanical machines can exhibit intelligence~\citep{DBLP:journals/ijssci/Wang09,Stradner2013coupled-collective-systems}---cf. Braitenberg vehicles~\citep{Dvoretskii2022braitenberg}, some of which are purely mechanical vehicles with hard-wired connections between sensors and actuators.
}
Common sub-fields of \ac{aci}
 include e.g. semantic web, social networks, and multi-agent systems~\citep{DBLP:conf/iccci/2009}.
Often, \ac{aci} and \ac{cci}
 include systems comprising both machines and humans.
Possible taxonomies for \ac{aci} are proposed in the next section.

Notice that terms like swarm intelligence
 or multi-agent intelligence
 may refer to natural, to artificial systems, 
 or to a general model comprising both.

Other kinds of \ac{ci} are described in the following,
 as they are very much related to 
 peculiar \ac{ci} engineering methods and techniques.

\section{Perspectives of \revA{Artificial} Collective Intelligence Engineering}
\label{sec:ci-eng}

\revA{Building on the previous discussion of \emph{what} \ac{ci} is and its main models,} this section focusses on \emph{how \ac{ci} can be engineered}, according to literature,
hence addressing \ref{rq-eng} to \ref{rq-field}.
\revA{In doing this, we will picture a map of the state of the art in \ac{ci} engineering, setting the stage for a discussion of research opportunities and challenges in the next section.}



Depending on what kind of \ac{ci} has to be achieved (cf. the previous section), 
 various \emph{perspectives} and approaches to \emph{\ac{ci} engineering}
 can be devised,
 each one leveraging and providing peculiar sets of \emph{\ac{ci} mechanisms}.
%

%
%

\subsection{Knowledge-oriented vs. behaviour-oriented \acs{ci}}
\label{ssec:knowledge-vs-behavior}%

From an industrial point of view, 
 the engineering of \ac{ci}
 often revolves around engineering the \ac{ict} platforms and algorithms
 for collecting data from human activity and extracting knowledge from collected data~\citep{segaran2007programming-ci,alag2008ci-in-action}.
There are several ways in which humans using web applications can provide data through their interaction:
 e.g., by what content they search, 
 what paths they follow,
 what feedback they provide,
 what content they add, etc.
Then, techniques like data mining, text mining, and machine learning can be used to classify information, cluster information, predict trends, recommend content, filter information, aggregate information etc.
We may call this \revB{\emph{knowledge-oriented}} \ac{ci}
 since the collective intelligence
 lies in the data produced \revB{and processed} by a collective,
 and \ac{ict} has a role in supporting such \revB{information creation, ultimately promoting the emergence of} \emph{latent} collective knowledge.
This is essentially what \cite{surowiecki2005wisdom-crowds}
 call\revA{s} \emph{cognition} problems.

\revA{
This kind of systems may not seem a form of \ac{ci}.
Indeed, one might be tempted to completely abstract over the collective of agents providing the data, and merely consider a conceptually single source of data and how data is processed and aggregated by a conceptually single process. 
However, the \ac{ci} nature of all this starts to emerge
 once one considers the overall process by a larger, socio-technical perspective.
By this perspective, several agents produce information through their activity and reasoning, possibly interacting with other agents and with supporting tools (e.g., a network to share information, tools to make sense of others' contributions, etc.)---cf. the isolation, collaboration, and feedback-based paradigms. 
%
%
}

Conversely, we may call \emph{behaviour-oriented} \ac{ci}
 the collective intelligence
 that drives the global behaviour of a system.
This includes what \cite{surowiecki2005wisdom-crowds}
 refers to as \emph{coordination} and \emph{cooperation} problems.
Examples include 
 the form of intelligence
 driving the way in which robotic swarms move~\citep{navarro-2013-survey-collective-movement-robots},
 a computational ecosystem that self-organises into activity and communication structures~\citep{DBLP:journals/fgcs/PianiniCVN21},
 and a market that self-regulates itself~\citep{lo2015markets}.
However, as the latter example shows,
 since collective action 
 is connected with collective decision making,
 which in turn is connected to collective knowledge,
 the border between knowledge-oriented and behaviour-oriented \ac{ci} \revA{is} fuzzy
 and \revA{so these types of \ac{ci} should} not be thought as containers for mechanisms
 but rather as containers for typical \ac{ci} goals.

\revB{
The distinction between ``plain'' systems and \acp{sos}~\citep{DBLP:journals/csur/NielsenLFWP15}, e.g., based on the properties of autonomy, belonging, connectivity, diversity, and emergence~\citep{DBLP:conf/sysose/BoardmanS06},
 is also relevant in this discussion~\citep{DBLP:journals/ais/PeetersDBBNSR21}.
Research under the knowledge-oriented \ac{ci} umbrella, typically involving socio-technical systems, seems mostly related to the \ac{sos} framework.
Research under the behavior-oriented \ac{ci} umbrella, instead,
 seems more uniformly distributed
 along both the system (cf. swarm intelligence and aggregate computing---\cite{bonabeau1999swarmintelligence-book,DBLP:journals/jlap/ViroliBDACP19})
 and \ac{sos} viewpoints (cf. hybrid human-machine systems---\cite{DBLP:journals/ais/PeetersDBBNSR21,DBLP:journals/tetc/ScekicSVRTMD20}).
}
 
\subsection{Manual vs. automatic \acs{aci} development}
\label{ssec:manual-vs-automatic}
Regarding \ac{aci},
 it is possible to distinguish two main kinds of approaches:
 those based on \emph{manual design} and those based on \emph{automatic design}.

\subsubsection{Manual design of \acs{aci}}\label{sssec:manual-design}
In the manual approach, a designer 
 specifies the behaviour of the computational agents 
 making up the collective
 directly by providing \emph{behavioural rules} (or \emph{policies}).
%

%
Here, the key issue is determining what individual behaviour, 
 when replicated or combined with other behaviours
 or phenomena,
 can give rise to the desired emergent behaviour.
Programming approaches that are thought 
 to somehow address this goal
 are often known as \emph{macro-programming}
 in the literature~\citep{DBLP:journals/corr/abs-2201-03473},
 a term that recurred especially in early 2000s
 in the context of \acp{wsn}~\citep{DBLP:conf/ipsn/NewtonAW05}.
\revA{
A foundational contribution to macro-programming is given by~\cite{DBLP:journals/swarm/ReinaMDT15},
 where a methodology is proposed for passing 
 from macroscopic descriptions
 to a microscopic implementation
 through a design pattern,
 obtaining a quantitative correspondence between micro and macro dynamics.
Research has produced different macro-programming frameworks,
 e.g., 
 for expressing the behaviour of robot swarms~\citep{DBLP:journals/software/PinciroliB16}
 and distributed, \ac{iot} systems~\citep{DBLP:conf/iotdi/NoorTGS19,DBLP:conf/cgo/MizziEP18}---though most of them lack formal foundations.
} 
 
A notable macro-programming approach 
 that has recently been subject to intense research
 is \emph{\ac{ac}}~\citep{DBLP:journals/jlap/ViroliBDACP19}.
\ac{ac} consists of a 
 functional macro-programming language
 expressing collective behaviour 
 in terms of computations over
 distributed data structures called \emph{computational fields}~\citep{DBLP:journals/pervasive/MameiZL04,DBLP:journals/jlap/ViroliBDACP19}.
The basic language constructs \revA{provide support for dealing} with 
 (i) lifting standard values to field values;
 (ii) abstracting field computations through functions;
 (iii) stateful evolution of fields; and
 (iv) handling bi-directional communication through so-called \emph{neighbouring fields}.
Using such constructs, 
 and library functions e.g. handling information flows through gradients or supporting higher-level patterns,
 a programmer can write an aggregate program
 that expresses the global behaviour 
 of a possibly dynamic network of agents.
The agents, by repeatedly evaluating the program 
 in asynchronous sense-compute-interact rounds,
 and interacting with neighbours by exchanging data 
 as dictated by the program,
 could steer self-organising behaviour 
 hopefully fulfilling the intent of the program.
\cite{DBLP:journals/eaai/CasadeiVAPD21} 
 argue that multiple concurrent and dynamic aggregate computations
 could pave a path to \ac{ci} engineering.

While \ac{ac} 
 adopts a swarm-like self-organisation model,
 another class of approaches for \ac{aci}
 is given by \emph{multi-agent programming},
 as supported e.g. by the JaCaMo platform~\citep{boissier2020multi-agent-oriented-prog-mitpress}, which comprises \emph{Jason} for programming cognitive autonomous agents, \emph{CArtAgO} for programming the distributed artifacts-based environment of the \ac{mas}, and \emph{MOise} for programming agent organisations.
However, it is worth noticing that
 the relationships between \ac{mas} research and \ac{ci} research
 are often hindered by different terminologies
 and separate communities.
A reason might be that a large part of \ac{mas} research properly 
 focusses on composites rather than collectives,
 i.e., well-structured organisations of heterogeneous intelligent agents
 rather than self-organising swarms of largely homogeneous and cognitively simple agents.

\subsubsection{Automatic design of \ac{aci}}\label{sssec:auto-design}
Since manually crafting control and behavioural rules
 of computational agents
 might be difficult, 
 especially for complex tasks
 in non-stationary environments,
 a different approach consists
 in devising strategies for automatically designing behaviours.
\revA{
The idea is to provide hints about the intended behaviour or the results to be attained by it
(e.g., in terms of \emph{specifications} or \emph{data}),
 and to leverage mechanisms to 
 generate or find behaviours that satisfy the specification.
}
This can be addressed through \revA{\emph{automatic programming}~\citep{DBLP:journals/gpem/ONeillS20},
 \emph{(machine) learning}~\citep{DBLP:conf/mrs/BehjatMKJCGDDDE21},
 and 
 \emph{search}~\citep{DBLP:books/aw/RN2020}.
For \ac{ci} systems, these are essentially the approaches followed by prominent methods like, e.g., 
 \acf{marl}~\citep{DBLP:journals/tsmc/BusoniuBS08}
 and evolutionary swarm robotics~\citep{DBLP:series/sci/2008-108}}.

One of the early models and notable example is \emph{COllective INtelligence}, or \emph{COIN}~\citep{wolpert2003ci}.
Essentially, COIN considers a \emph{collective} as a system of self-interested agents, trying to maximise their \emph{private utility} function,
 sharing an associated \emph{world utility}
 giving a measure of the \ac{ci} of the overall system.
\ac{marl} is clearly a powerful technique for building \ac{ci}, and it is currently a hot research area,
 with several surveys emerging~\citep{DBLP:journals/corr/abs-1911-10635,canese2021marl-review-challenges-apps,DBLP:journals/air/GronauerD22}.
\revA{
Learning of collective behaviour may be related but should not be confused with collective learning, which is learning carried out by multiple agents that does not necessarily yield collective behaviour models. 
}

\revA{
In evolutionary robotics~\citep{DBLP:series/sci/2008-108},
 the idea is to use evolutionary algorithms
 (i.e., algorithms that use mechanisms inspired by biological evolution for evolving populations of solutions)
 to optimise models of robot controllers 
 (e.g. mapping inputs from sensors to outputs to actuators)
 with respect to desired behavioural goals.
Various techniques have been proposed in the literature to improve traditional evolutionary approaches, e.g., novelty search~\citep{DBLP:journals/swarm/GomesUC13}.
An interesting approach for the automatic design of the control logic of swarms is given by \emph{AutoMoDe}~\citep{DBLP:journals/swarm/FrancescaBBTB14}.
AutoMoDe generates modular control software
 as a probabilistic finite state machine by selecting, composing, and configuring behavioural modules (bias).
The idea is to leverage the bias to make the automatic design approach robust to differences between simulation and reality.
%
%
Another relevant methodology for evolutionary robotics is the so-called \emph{embodied evolution} approach~\citep{DBLP:journals/ras/WatsonFP02,DBLP:journals/firai/BredecheHP18}, 
 which is based on evolutionary processes 
 that are \emph{distributed }
 in a population of robots
 situated in an environment,
 to support online and long-term adaptivity.
Embodied evolution is an interesting setting
 for studying aspects like embodied intelligence,
 co-evolution, the role of environmental niches, the relationship between optimisation and selection pressure, locality of interaction, etc.
The combination of learning and evolution 
 is also a very interesting research direction~\citep{Gupta2021embodied-intelligence-learning-evolution}.
}

A second possibility for automatic design
 comes from \emph{program synthesis}~\citep{DBLP:journals/ftpl/GulwaniPS17},
 which is the field studying the task of automatically crafting programs (in some given programming language) that satisfy a specified intent.
Particularly interesting are the recent attempts of 
 combining program synthesis and reinforcement learning---cf.~\cite{DBLP:conf/icml/BastaniIS20,aguzzi2022coord-ac-rl}.
However, in the context of \ac{ci}, this direction has not yet been investigated, representing an opportunity for future research (cf. next sections).

\revA{
As a final remark,
 we stress that 
 manual and automatic design
 can be seen as the extremes of a continuum,
 and that hybrid approaches can be used---cf. interactive program synthesis~\citep{DBLP:conf/uist/0001LWG20}.
}

\subsection{Relationships between humans and machines in \acs{hmci}}
\label{ssec:human-machine-ci}
In \ac{hmci}, it is possible to distinguish multiple threads of research. 
A first classification could be based on the aforementioned distinction between knowledge-oriented and behaviour-oriented \ac{ci}.
Other classifications can be made by considering what kind of entity plays the role of \emph{controller} and \emph{executor}:
\begin{enumerate}
	\item tasking crowds of humans~\citep{DBLP:journals/cm/GantiYL11,DBLP:conf/percom/GuoYZZ14,DBLP:journals/smr/ZhenKNZAK21,DBLP:journals/jss/SariTA19}---cf. crowdsourcing~\citep{DBLP:journals/smr/ZhenKNZAK21} and crowdsensing~\citep{DBLP:conf/percom/GuoYZZ14};
	\item using humans to guide machine operations, e.g., interactively~\citep{DBLP:journals/ccftpci/YuLYG21};
\end{enumerate}
or considering what entity plays the role of \emph{input} and \emph{output}
\begin{enumerate}
\item using \ac{ai} to extract or mine intelligence from human contributions~\citep{segaran2007programming-ci,alag2008ci-in-action};
\item using humans (or \emph{human computation}) to extract value from machine contributions, especially in tasks where machines cannot (yet) generally perform well, such as visual recognition and language understanding~\citep{DBLP:conf/chi/QuinnB11}.
\end{enumerate}
or, finally, considering humans and machines as peers and hence the so-called
\emph{human-machine networks}~\citep{DBLP:journals/csur/TsvetkovaYMPEWL17}
	 or \emph{social machines}~\citep{DBLP:conf/www/BuregioMR13,berners1999weaving-the-web}.

Regarding the engineering of social machines,
 a notable macro-programming approach
 is given by the \emph{SmartSociety} platform~\citep{DBLP:journals/tetc/ScekicSVRTMD20},
 which is based on 
 abstractions like
 persistent and transient teams of human/machine peers,
 and collective-based tasks.
The approach can be used for human orchestration 
 and human tasking activities like those found in crowdsourcing and hybrid collectives.

\revB{
Concerning the general design of \ac{aci} in social machines, \cite{DBLP:journals/ais/PeetersDBBNSR21} propose three principles:
 (i) goals from the collective, technological, and human perspectives should be considered simultaneously;
 (ii) development effort should continously embrace all the product's lifecycle;
 and
 (iii) the requirements of observability, predictability, explainability, and directability should be considered at all abstractions levels (\ac{ai}, team, and society).
}


\revA{
\subsection{Collective tasks}
\label{ssec:collective-tasks}
Another main classification of \ac{ci} engineering research is by the kind of collective task that is addressed.
A \emph{collective task} can be defined as a task
 that \emph{requires} more than one individual
 to be carried out.
Notice that \ac{ci} may be seen as a requirement or mechanism for solving collective tasks (cf. the general \ac{ci} interpretation)
 or, conversely,
 \ac{ci} might be defined (and measured) in terms of the ability to solve a set of collective tasks in a variety of environments.

Multiple taxonomies of collective tasks have been proposed in the literature.
For instance, \cite{DBLP:journals/swarm/BrambillaFBD13}
  classify collective behaviours (of swarm robotics systems)
  into
  (i) spatially-organising behaviours,
  (ii) navigation behaviours,
  (iii) collective decision making,
  and
  (iv) others.
Other reviews of swarm robotics tasks include~\citep{DBLP:journals/ijon/Bayindir16,DBLP:journals/swevo/NedjahJ19}.
Moreover,
 collective tasks can be classified also
 according the three paradigms discussed in~\citep{he2019ci-taxonomy-survey} and reviewed in previous sections: isolation, collaboration, and feedback.

In the following, we review material 
 for two general, main kinds of collective tasks --
 collective decision making and collective learning --
 and then point out references to other kinds of tasks.

\subsubsection{Collective decision making}
\label{ssec:collective-decision-making}
Collective decision making
 is the problem of how groups reach decisions without any centralised leadership~\citep{Bose2017collective-decision-making,DBLP:journals/swarm/PrasetyoMF19}.
This is also known as \emph{group decision making}~\citep{DBLP:journals/eor/ZhangDCY19,Tang2021group-decision-making}.

Decision making and its collective counterpart can be 
 classified according to the nature of the decision to be made.
Reaching consensus and multi-agent task allocation are two common kinds of collective decision-making behaviours, typical in swarm robotics~\citep{DBLP:journals/swarm/BrambillaFBD13}.
\cite{DBLP:conf/mates/Guttmann09}
 classifies \ac{mas} decision making by four dimensions:
 \emph{(i) use of models of self vs. models of others};
 \emph{(ii) individual inputs vs. group input};
 \emph{(iii) learning vs. non-learning}, depending on whether decision making spans multiple rounds or just one round;
 and
 \emph{(iv) collaboration vs. competition}.
\cite{surowiecki2005wisdom-crowds}
 distinguishes three kinds of problems or tasks
 of distributed decision making:
 \emph{(i) cognition}, 
 \emph{(ii) cooperation},
 and \emph{(iii) coordination}.

Collective decision making is often supported by self-organisation mechanisms \revB{based on, e.g., collective perception~\citep{DBLP:conf/sab/SchmicklMC06},} voter models~\citep{DBLP:conf/atal/ValentiniHD14}, opinion formation models~\citep{DBLP:journals/swarm/OcaFSPBD11}, and self-stabilising leader election~\citep{Pianini2022self-stab-leader-election}.

Recent surveys on collective decision making include the following.
\revB{\cite{DBLP:journals/firai/ValentiniFD17}
 focus on discrete consensus achievement, and 
 propose a formal definition of the \emph{best-of-$n$} problem (choice of the best alternative among $n$ available options);
 then, they 
 define a taxonomy based on different classes of the problem, 
 and classify the literature on discrete consensus agreement accordingly.
%
}
\cite{DBLP:journals/eor/ZhangDCY19}
 provide a review of consensus models in collective decision making, and compare them based on multiple criteria for measuring consensus efficiency.
They also argue that two interesting research directions include (i) \emph{large-scale} collective decision making
and (ii) addressing social relationships and opinion evolution.
\cite{Tang2021group-decision-making}
 provide a review of literature around five challenges in large-scale collective decision making with big data:
 dimension reduction,
 weighting and aggregation of decision information,
 behaviour management,
 cost management,
 and knowledge distribution and increase.
\cite{DBLP:journals/tamd/RizkAT18}
 provide a survey of decision making in \acp{mas}.
The survey focusses on five cooperative decision-making models: Markov decision processes (and variants), control theory, game theory, graph theory, and swarm intelligence.
These models are discussed along the dimensions of heterogeneity, scalability, and communication bandwidth---which are also crucial research challenges.
Particularly challenging is also decision making in dynamic environments~\citep{DBLP:journals/tamd/RizkAT18,DBLP:journals/swarm/PrasetyoMF19}.
Other challenges include security, privacy, and trust;
 approaches to address these include, e.g., blockchain consensus~\citep{DBLP:journals/jpdc/Pournaras20}.

\subsubsection{Collective learning}
\label{ssec:collective-learning}
Learning is intimately related to intelligence~\citep{Jensen1989-learning-intelligence-relationship}.
Collective learning is learning backed by a collective process,
 with coordination and exchange of information between individuals and artifacts~\citep{Fadul2009collective-learning}.
As a multi-disciplinary theme, it is studied both in areas like sociology and organisational theory~\citep{Garavan2012collective-learning,Fadul2009collective-learning},
 and in \ac{ai} research~\citep{bock1993emergence-collective-learning}.
Collective learning spans both the knowledge-oriented and behaviour-oriented perspectives of \ac{ci},
 and is the main technique for automatic design of \ac{aci}.
Goals of collective learning 
 include 
 supporting individual learning~\citep{Fenwick2008individual-vs-collective-learning},
 producing collective knowledge,
 and promoting collective decision making~\citep{Garavan2012collective-learning}.
As a wide concept,
 collective learning can be interpreted along multiple perspectives~\citep{Garavan2012collective-learning}: e.g., as the independent aggregation of individual learning, or as a collaborative activity.
So, collective learning is related but not necessarily the same as cooperative and collaborative learning~\citep{Fadul2009collective-learning}.
These different views can also be found in \ac{ai} and \ac{aci} research.

Artificial collective learning includes distributed machine learning~\citep{DBLP:journals/csur/VerbraekenWKKVR20}:
 examples include centralised, federated, and decentralised machine learning systems.
In \emph{centralised learning},
 the different individuals of the system
 provide data to a central entity that performs the actual learning process.
So, in this case, the core learning process is not collective, though it would be collective if considered by a larger perspective that includes data generation.
In \emph{federated learning}~\citep{DBLP:journals/ftml/KairouzMABBBBCC21},
 the idea is that individual independent workers perform machine learning tasks on local data sets, producing models that are then aggregated by a master into a global model without the need of sharing data samples. It enables to address data privacy issues. The combination of multiple models is also called \emph{ensemble learning}~\citep{DBLP:journals/fcsc/DongYCSM20}.
\cite{DBLP:journals/jpdc/HegedusDJ21}
 propose \emph{gossip-based learning} 
 as a decentralised alternative to \emph{federated learning},
 where no central entities are used
 and models are gossiped and merged throughout the nodes of the system.
Collective learning might be supervised or unsupervised.
An example of an unsupervised decentralised collective learning approach is provided by~\cite{DBLP:journals/taas/PournarasPA18}.

Another important example of collective learning is \ac{marl}~\citep{DBLP:journals/tsmc/BusoniuBS08},
 which considers learning by collections of reinforcement-learning agents.
\ac{marl} algorithms are commonly classified depending on whether they address \emph{fully cooperative}, \emph{fully competitive}, and \emph{mixed cooperative/competitive} problems.
In fully cooperative problems,
 the agents are given a \emph{common reward signal} that evaluates the collective action of the \ac{mas}. 
Instead, in fully competitive problems,
 the agents have opposite goals.
Mixed games are in between fully cooperative and fully competitive problems.
Three common information structures in \ac{marl} are~\citep{DBLP:journals/corr/abs-1911-10635}: 
\emph{(i) centralised structures}, involving a central controller aggregating information from the agents;
\emph{(ii) decentralised structures}, with no central entities and neighbourhood interaction;
and
\emph{(iii) fully decentralised}, namely independent learning, with no information exchanged between the agents.
Various formal frameworks have been proposed to address  \ac{marl} problems,
including \emph{COIN}~\citep{wolpert2003ci} and \emph{Decentralised Markov Decision Processes (Dec-MDP)}~\citep{oliehoek2016intro-dec-pomdp}.
The reader interested to \ac{marl} algorithms and frameworks can check out multiple comprehensive surveys on the topic~\citep{DBLP:journals/tsmc/BusoniuBS08,DBLP:journals/corr/abs-1911-10635,DBLP:journals/aamas/Hernandez-LealK19}.

There exist surveys on collective learning.
\cite{DBLP:conf/icse/DAngeloGGGNPT19} 
 perform a systematic literature review on learning-based collective self-adaptive systems.
Their analysis extracts, as the main characteristics of such systems, 
 the application domains involving groups of agents with the ability to learn,
 the levels of autonomy of the agents,
 the levels of knowledge access (i.e., the way in which they explicitly share learning information),
 and the kinds of behaviours involved (e.g., selfish vs. collaborative).
Accordingly, the authors provide a framework for learning collective self-adaptive systems, based on three dimensions: autonomy, knowledge access, and behaviour.
The learning goals 
 are analysed w.r.t. the target emergent behaviour;
 from the analysis, two clusters of works emerge:
 those where the emergent behaviour is associated to the anticipated learning task, and those where it is not.
Among the learning techniques, 
 they report 
 that the majority of research works
 leverage reinforcement learning,
 while game theory, supervised learning, probabilistic and other approaches are less investigated in these settings.
Resilience and security are deduced as the main open challenges in this research domain. 

\cite{DBLP:conf/acsos/Pournaras20} provides a review of 10-years research on human-centred collective learning for coordinated multi-objective decision making in socio-technical systems, within the context of the \emph{Economic Planning and Optimized Selections (EPOS)} project. 
Collective learning is motivated as a way to address the long-standing \emph{tragedy of the commons} problem,
and argued to be a promising paradigm of artificial intelligence.
As research opportunities and challenges, the author identifies: 
explainability and trust, 
resilience to plan violations and adversaries,
collective learning in organic computing systems,
co-evolution of collective human and machine learning,
and digital democracy.

\revB{Learning is also very related to evolution~\citep{DBLP:journals/firai/BredecheHP18}.
Learning and evolution are generally considered as different mechanisms for adaptation 
 working on different time and spatial scales~\citep{DBLP:books/daglib/0024632,anderson2013adaptive}.
However, these techniques can also be combined~\citep{DBLP:journals/arobots/NolfiF99}:
 learning can guide evolution~\citep{DBLP:journals/compsys/HintonN87} and evolution can improve learning (cf. evolutionary learning---\cite{DBLP:journals/csur/TelikaniTBG22}),
 where different architectures for the combination are possible~\citep{Sigaud2022evolution-deeprl}.
}

\subsubsection{Other collective tasks}
\label{ssec:collective-tasks-other}
\emph{Collective action}~\citep{Oliver1993-formal-models-collective-action}
 commonly refers to the situation where multiple individuals
 with conflicting goals as well as common goals
 would benefit from coordinated action.
Clearly, the ability to act collectively in an effective manner can be seen as an expression of \ac{ci}.
The problem is addressed mainly in sociology,
 but computer science also provides tools (e.g., simulations, models etc.),
 such as the \emph{SOSIEL (Self-Organising Social \& Inductive Evolutionary Learning)} simulation platform~\citep{Sotnik2018sosiel-platform}, for studying the problem and investigating solutions 
 for human societies as well as for socio-technical and artificial systems.
Collective actions may be supported by collective and self-organised decision-making processes, and leveraging abstractions like \emph{electronic institutions} and \emph{social capital}~\citep{DBLP:conf/saso/PetruzziBP15}.

\emph{Collective movement}~\citep{navarro-2013-survey-collective-movement-robots} is the problem of making a group of agents (e.g., robots, drones, vehicles) move towards a common direction in a cohesive manner.
Notice that this is not just about movement per se,
 but rather moving in conjunction or in order to support other tasks as well---e.g., distributed sensing, exploration, and rescue tasks.
Two main sub-problems can be identified~\citep{navarro-2013-survey-collective-movement-robots}:
 (i) \emph{formation control}~\citep{DBLP:journals/comsur/YangXL21}, when the shape of the group and/or the individual positions' are important;
 and
 (ii) \emph{flocking}~\citep{DBLP:journals/arc/BeaverM21}, where such aspects are less important.

\emph{Distributed optimisation}~\citep{DBLP:journals/arc/YangYWYWMHWLJ19}
 refers to the problem
 of minimising a global objective function,
 which is the sum of the local objective functions of the members of a collective,
 in a distributed manner.
Distributed optimisation can be a technique for collective decision making.

\emph{Collective knowledge construction} refers to the creation of new, distributed, and shared knowledge by a collective~\citep{Hecker2012-collective-knowledge}. 
This topic is generally studied by considering aspects such as
 collaboration~\citep{DBLP:journals/ce/Hmelo-Silver03},
 socio-technical infrastructures~\citep{DBLP:journals/ws/Gruber08},
 knowledge transfer~\citep{Huang2018collective-knowledge-transfer},
 the interplay between individual and collective knowledge~\citep{Kimmerle2010interplay-individual-coll-k}
 models of information diffusion dynamics~\citep{DBLP:journals/eis/Maleszka19},
 and lifelong learning~\citep{DBLP:conf/atal/RostamiKKE18}.
}

\subsection{A view of \acs{ci}-related fields}
\label{ssec:related-fields}

Being \ac{ci} a multi-disciplinary field,
 the engineering of \ac{ci} and \ac{aci}
 can benefit from ideas and research results 
 from a variety of fields.
It would be useful to have a comprehensive map
 of research fields contributing to \ac{ci}.

\begin{figure*}
\centering
\includegraphics[width=0.8\textwidth]{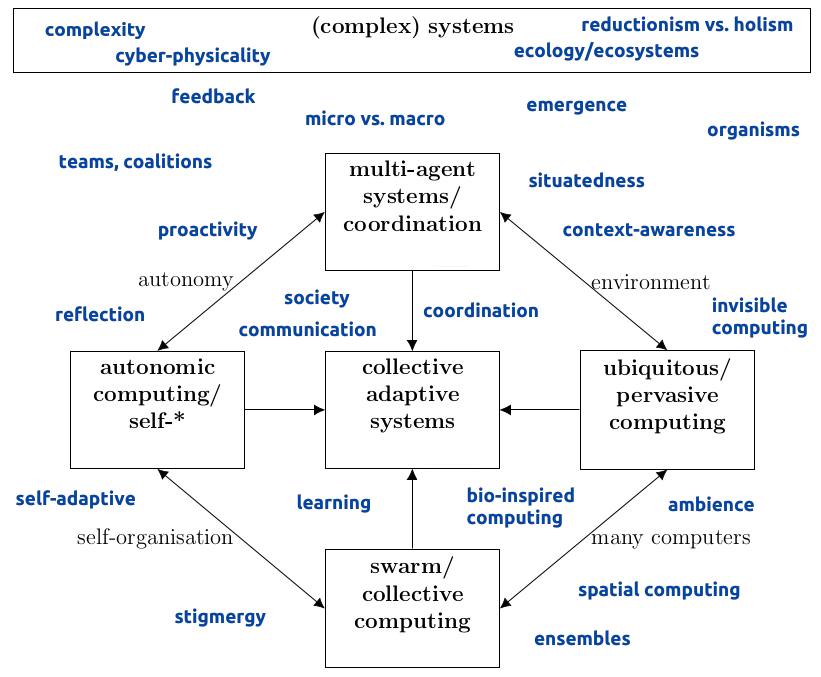}
\caption{A research map of fields and concepts contributing to (research on) \ac{ci} engineering.}
\label{fig:cas-fields}
\end{figure*}

Though we consider providing a comprehensive research map of \ac{ci} engineering
 as a future work,
 we provide a research map
 (see \Cref{fig:cas-fields})
 from the perspective of 
 \emph{\ac{cas}} research~\citep{casadei2020phd-thesis,DBLP:journals/sttt/NicolaJW20,DBLP:journals/tasm/BucchiaroneDPCS20,DBLP:conf/huc/Ferscha15}.
The idea is that 
 \ac{ci} engineering
 should be supported 
 through inter-disciplinary research
 and 
 a systems science perspective~\citep{mobus2015principles-systems-science},
 also providing a rigorous treatment 
 of system-level properties
 that could be sustained by \ac{ci} processes.
This includes leveraging studies 
 of abstract and fundamental kinds of systems
 such as, for instance, \ac{cps}, namely systems that combine discrete and continuous dynamics~\citep{alur2015principles-cps}.
Then, a 
 set of inter-related fields can 
 promote the study of peculiar \ac{ci} phenomena
 such as emergence, self-organisation, ensemble formation, etc.
Such fields include but are not limited to the field of 
coordination~\citep{DBLP:journals/csur/MaloneC94},
multi-agent systems~\citep{DBLP:books/daglib/0023784},
autonomic/\selfstar{} computing~\citep{DBLP:journals/computer/KephartC03},
collective adaptive systems~\citep{casadei2020phd-thesis,DBLP:journals/sttt/NicolaJW20,DBLP:journals/tasm/BucchiaroneDPCS20,DBLP:conf/huc/Ferscha15},
ubiquitous/pervasive computing~\citep{weiser1991computer},
swarm intelligence~\citep{bonabeau1999swarmintelligence-book},
and collective computing~\citep{DBLP:journals/computer/Abowd16}.
\revA{Some of these are briefly overviewed in the following.}

\revA{
We noticed multiple times in previous sections how interaction is a key element of \ac{ci}.
\emph{Coordination} is the interdisciplinary study of interaction~\citep{DBLP:journals/csur/MaloneC94}.
In computer science, 
 interaction was early recognised as a concern 
 related but clearly distinguished from computation~\citep{DBLP:journals/cacm/GelernterC92},
 hence amenable to separate modelling by so-called coordination languages.
A general meta-model of coordination~\citep{DBLP:journals/csur/Ciancarini96} 
 consists of \emph{coordinables} (the interacting entities),
 \emph{coordination media} (the abstractions supporting and constraining interactions),
 and 
 \emph{coordination laws} (describing the behaviour of a coordination medium).
Languages, abstractions, and patterns, 
 can be used to define 
 the way in which computational components coordinate 
 across aspects like 
 control, information, space, and time.
This has motivated the birth of whole communities and long-standing research threads~\citep{DBLP:conf/coordination/1996,DBLP:conf/coordination/2022}.

\emph{\Acfp{cas}}
 are collectives of agents
 that can adapt to changing environments
 with no central controller.
Their engineering poses several challenges, tackled in corresponding research communities~\citep{DBLP:journals/sttt/NicolaJW20,DBLP:journals/tasm/BucchiaroneDPCS20}.
\acp{cas} are sometimes considered to be heterogeneous~\citep{DBLP:conf/sfm/LoretiH16,DBLP:conf/icsoc/AndrikopoulosBSKM13}, contrasted to more homogeneous intelligent swarms, though we tend to disagree with this view.
In our view, \acp{cas} are a superset of intelligent swarms,
 which are characterised by \emph{(i) large numbers of (ii) relatively simple (or not particularly intelligent) individuals}~\citep{DBLP:conf/sab/Beni04}.
Collectives are generally \emph{quite homogeneous, at least at some level of abstraction}~\citep{DBLP:journals/taas/PianiniPCE22}, 
 though research works aim to address heterogeneous collective adaptive systems~\citep{DBLP:journals/tetc/ScekicSVRTMD20} as well as heterogeneous swarms~\citep{DBLP:journals/ram/DorigoFGMNBBBBBBCCDCDFFGGLMMOOPPRRSSSSTTTV13,DBLP:conf/prima/KengyelHZRWS15}, e.g., with systems involving humans and robots~\citep{DBLP:journals/adb/HasbachB22}, or groups of robots with different morphology or behaviour.
Swarm robotics is the combination of swarm intelligence and robotics~\citep{DBLP:journals/swarm/BrambillaFBD13,DBLP:conf/sab/Beni04}.

Coordination, \acp{cas}, and swarm robotics 
 can also be seen as sub-fields
 of the larger field of \acp{mas}~\citep{DBLP:books/daglib/0023784,DBLP:conf/atal/2022},
 which itself stemmed from the field of distributed artificial intelligence~\citep{DBLP:books/daglib/0000780}.
In \acp{mas} engineering, two main levels and corresponding problems are considered:
 the \emph{micro level} of agent design,
 and the \emph{macro level} of agent society design.
\emph{Autonomy} (encapsulation of control) 
and \emph{agency} (the ability to \emph{act}) are generally considered the 
 two fundamental properties of agents~\citep{DBLP:conf/atal/FranklinG97,DBLP:journals/igpl/HoekW03a},
 from which other properties like 
 proactiveness, interactivity, and sociality stem.
By a software programming and engineering point of view,
 agents can be considered as an abstraction 
 following active objects and actors~\citep{DBLP:journals/jot/Odell02} that,
 together other first-class abstractions
 like artifacts~\citep{DBLP:journals/aamas/OmiciniRV08}, environments~\citep{DBLP:conf/e4mas/WeynsM14}, and organisations~\citep{DBLP:journals/ker/HorlingL04},
 provide a support for the so-called \emph{(multi-)agent-oriented programming} paradigm~\citep{DBLP:journals/ai/Shoham93,boissier2020maop}.
The \ac{mas} field/perspective is clearly intimately related to \ac{ci}.

Like for \acp{mas},
 the key property of autonomy
 is at the centre of \emph{autonomic computing}~\citep{DBLP:journals/computer/KephartC03},
 namely the field devoted to the construction of
 computational systems that are able to manage/adapt themselves
 with limited or no human intervention.
Following this vision, 
 research has been carried out to find 
 approaches and techniques 
 to endow artificial systems 
 with different \emph{\selfstar{} properties}:
 self-adaptive~\citep{DBLP:conf/dagstuhl/LemosGMSALSTVVWBBBBCDDEGGGGIKKLMMMMMNPPSSSSTWW10,DBLP:journals/taas/SalehieT09},
 self-healing/repairing~\citep{DBLP:journals/computing/PsaierD11},
 self-improving/optimising~\citep{DBLP:conf/saso/BellmanBDEGLLST18},
 self-organising~\citep{DBLP:reference/sp/Heylighen13}, and so on.
To build autonomic systems, 
 approaches typically distinguish 
 between the \emph{managed system}
 and the \emph{managing system},
  structuring the latter
  in terms of 
  \emph{Monitoring, Adaptation, Planning, Execution, and Knowledge (MAPE-K)} components~\citep{DBLP:journals/computer/KephartC03}.
In so-called \emph{architecture-based self-adaptation}~\citep{DBLP:journals/computer/GarlanCHSS04},
 architectural models of the managed systems 
 are leveraged at runtime to organise the self-managing logic.
The managing system could also be distributed and decentralised~\citep{DBLP:conf/dagstuhl/WeynsSGMMPWAGG10}.
If the managed system is a collective,
 then its \selfstar{} properties could be put in relation to its \ac{ci}.
Consider the property of being \emph{self-organising}, 
 characterised by processes
 that autonomously and resiliently increase/maintain order or structure~\citep{DBLP:conf/atal/WolfH04}; it typically emerges from the interaction of several entities.
Self-organisation can be considered as a key promoter or element of \ac{ci}~\citep{DBLP:journals/advcs/RodriguezGR07}.

As a last remark,
 we stress that the aforementioned fields 
 are highly inter- and trans-disciplinary.
For instance, \acp{mas} can be considered 
 by economical, sociological, organisational, and computational perspectives~\citep{DBLP:books/daglib/0023784}.
Same goes for coordination~\citep{DBLP:journals/csur/MaloneC94}.
Moreover, a great source of inspiration is given 
 by natural (e.g., physical and biological) systems,
 as recognised by 
 a wealth of 
 \emph{nature-inspired coordination}~\citep{DBLP:journals/percom/ZambonelliOACAS15} and 
 \emph{nature-inspired computing}~\citep{DBLP:journals/cogcom/SiddiqueA15} contributions.
}

%
%
%
%
%

\section{Research Opportunities and Challenges}
\label{sec:research-challenges}
\revA{With an understanding of the nature of \ac{ci} and its  engineering perspectives,} in the following,
 we discuss a few related research directions 
 that include interesting opportunities and challenges
 for researchers in \ac{ci} engineering.

\subsection{Programming emergence and macro-programming}
The problem of programming emergent and self-organising behaviours is an open research challenge~\citep{DBLP:journals/alife/GershensonTWS20,varenne2015programming-emergence}
 intimately related to \ac{ci} engineering.
Term \emph{macro-programming} emerged in early 2000s 
 to identify programming approaches 
 with the goal of defining the global behaviour
 of \acp{wsn}~\citep{DBLP:conf/ipsn/NewtonAW05};
 currently, it generally denotes paradigms
 aiming at supporting the programming 
 of system- or macro-level properties and behaviours.
A recent survey by \cite{DBLP:journals/corr/abs-2201-03473}
 shows that, beside the first wave 
 of research in the context of \acp{wsn},
 we are witnessing a renewed interest 
 in macro-programming
 fuelled by scenarios like
 the \acl{iot}, robot swarms, 
 and collective adaptive systems in general.
This is also very much related to spatial computing~\citep{beal2013organizing},
 as space is often a constraint, a means, or a goal in systems. 
 
The key challenge here is 
 determining what local behavioural rules of the individuals
 can promote the desired collective behaviour.
In particular, we can distinguish two problems~\citep{wolpert1999introduction-ci-survey}.
Given a set of individuals
 and the corresponding local behavioural rules,
 the \emph{local-to-global mapping problem} (or \emph{forward problem})
 is the problem of determining
 what global outcomes will be produced.
Conversely, the
\emph{global-to-local mapping problem}
(or \emph{inverse problem})
is the problem of determining what local behaviours
have produced the observed global outcomes.
In macro-programming,
 the latter problem turns into
 how to map a description of a global intent (macro-program)
 into local behavioural rules (micro-programs)~\citep{DBLP:journals/corr/abs-2201-03473}.

It has been shown that approaches like 
 aggregate computing~\citep{DBLP:journals/jlap/ViroliBDACP19}
 can support forms of self-organisation and \ac{ci}
 with macro-programs that can be encoded
 as compositions of functions 
 of reusable collective behaviours~\revA{\citep{DBLP:conf/ecoop/AudritoCDSV22}}.
This is promising, 
 but still little research has been devoted yet
 at investigating, systematising, and formalising 
 the principles, concepts, and mechanisms of macro-programming in general or specific settings~\citep{DBLP:journals/corr/abs-2201-03473}.


\subsection{Integration of manual and automatic \ac{ci} engineering methods}
In previous sections,
 we have discussed
 how \ac{ci}
 can be programmed manually (e.g., through macro-programming languages, or using traditional techniques to connect and extract knowledge from human activity)
 or automatically
 (e.g., via multi-agent reinforcement learning techniques or program synthesis).
Arguably, the two approaches could be combined
 to overcome their individual issues.
This is a still an unexplored research direction,
 but early works and ideas are emerging.

A first idea could be to use program synthesis~\citep{DBLP:journals/ftpl/GulwaniPS17}
 to synthesise macro-programs
 expressed in a macro-programming language~\citep{DBLP:journals/corr/abs-2201-03473}.
This could be coupled with simulation
 to verify how systems executing synthesised programs
 operate in various environments.
On one hand, since simulations may also be computationally-intensive,
 it might be necessary to limit simulation
 to few program candidates.
On the other hand, the problem of generating macro-programs 
 might be hard 
 especially if the space of possible programs 
 is very large.
Therefore, macro-programming languages admitting few primitives or combinators may be more suitable for this.

Additionally, there exist some recent attempts at 
 combining program synthesis and reinforcement learning~\citep{DBLP:conf/icml/VermaMSKC18,DBLP:conf/icml/BastaniIS20,qiu2022programmatic-rl-without-oracles}.
For instance, \cite{DBLP:conf/icml/BastaniIS20}
 discuss approaches to reinforcement learning
 based on learning programmatic policies (i.e., policies in the form of a program),
 which can provide benefits in terms of interpretability, formal verification, and robustness.
Therefore, it would be interesting 
 to consider the application of \ac{marl}
 where policies are expressed 
 in a multi-agent oriented
 or a macro-programming language.
An early attempt has been carried out e.g. in~\citep{aguzzi2022coord-ac-rl},
 where \ac{marl}
 has been used to 
 fill a hole
 in a sketched aggregate computing program
 (cf. the \emph{sketching} technique in program synthesis~\citep{DBLP:conf/aplas/Solar-Lezama09}),
 resulting in a collective adaptive behaviour
 that improves over a simple, manually encoded 
 collective behaviour.

\subsection{Integration of bottom-up and top-down processes}
Another interesting research challenge and opportunity
 for our ability of engineering \ac{ci} 
 lies in achieving a better understanding
 of how bottom-up and top-down processes
 can be integrated---or, in other words, how emergence
 and downward causation/feedback can be exploited altogether to provide both flexibility and robustness in collective behaviour.
Indeed,
 we are considering 
 \emph{feedback} \ac{ci} paradigm~\citep{he2019ci-taxonomy-survey},
 where the aggregation of contributions from the individuals and the environment in turn affects the individuals and the environment.
This is also what \cite{DBLP:conf/medes/LykourentzouVL09}
 call \emph{active} \ac{ci} systems,
 where collective behaviour is supported by the system level, which are contrasted from \emph{passive} \ac{ci} systems
 where no collective awareness or intentionality is present.

The problem of integrating top-down and bottom-up processes
 is indeed connected with the problem
 of \emph{controlling emergence},
 addressed in research fields
 like autonomic computing~\citep{DBLP:journals/computer/KephartC03}, with its \emph{MAPE-K (Monitor--Analyse--Plan--Execute with Knowledge)} loops,
 and \emph{organic computing}~\citep{DBLP:books/sp/MT2017},
 with \emph{observer-controller} architectures.
One issue is that emergence itself is a controversial concept,
 subject to philosophical and scientific investigation, and 
 often presented with definitions that hardly apply to systems engineering~\citep{DBLP:conf/atc/Muller-SchloerS06}.
Attempts to defining emergence 
 based on hierarchical system models
 and ontological approaches~\citep{gignoux2017emergence-graph-theory}
 may prove useful.
Initial, working classifications of emergence 
for reasoning in systems engineering may be based e.g.
on whether it is \emph{anticipated} or \emph{not anticipated}, and whether it is \emph{desirable} or \emph{undesirable}~\citep{DBLP:conf/isdevel/Iivari16}.

Some engineering techniques discussed in this section, such as macro-programming and \ac{marl}, could support the design of ``controlled emergence'' and, on the other way, a deeper understanding on emergence and its relationship with feedback could provide insights for mechanisms or the implementation of such techniques.
A macro-program, indeed, could be seen as 
 a top-down structure for emergent processes.
Also interesting in this respect are e.g. 
 formal studies carried out on \emph{self-stabilisation} of aggregate computations~\citep{Pianini2022self-stab-leader-election}, which guarantees that stable outputs are eventually achieved from stable inputs.

\subsection{Integrating humanity and technology: social machines}

A key subfield of \ac{ci} 
 that is still at its early days
 is \acf{hmci}~\citep{smirnov2019human-machine-ci},
 also known as \emph{hybrid \ac{ci}}~\citep{DBLP:journals/ais/PeetersDBBNSR21,moradi2019collective-hybrid-intelligence},
 or \emph{hybrid \acp{cas}}~\citep{DBLP:journals/tetc/ScekicSVRTMD20}.
In the systems we consider in this article, 
 we can identify two main domains~\citep{beal2013organizing}:
 (i) the domain of \emph{space-time}, which corresponds to physical environments and their evolution;
 and (ii) the domain of \emph{information}, 
 which evolves through computation.
Of course, these two domains interact, e.g., by measuring space-time to get associated information, and using information to manipulate space-time, through actuations.
Now, addressing the integration of humans and machines
 passes through the realisation
 that both kinds of individuals 
 can fully operate on those two domains.
That is, humans can be thought as computing machines (cf. the concept of \emph{human computation}~\citep{DBLP:conf/chi/QuinnB11}),
 and (computing) machines 
 can operate in the physical world
 (cf. the notion of \emph{\acl{cps}}~\citep{alur2015principles-cps}).
Indeed, various terms or buzzwords are emerging
 to denote systems where such integration of 
 humans, computation, and physical systems
 is present---cf., human \acp{cps}~\citep{DBLP:journals/jzusc/LiuW20}, human-in-the-loop \acp{cps}~\citep{DBLP:journals/computer/SchirnerECP13},
 and crowd computing~\citep{murray2010crowd-computing}.
From the perspective of computing,
 it is worth noting that
 \emph{collective computing}
 based on heterogeneous human-machine collectives
 was identified by \cite{DBLP:journals/computer/Abowd16}
 as the fourth generation in computing 
 following Weiser's characterisation of evolution of computing 
 from mainframe computing
 to personal computing
 to ubiquitous computing~\citep{weiser1991computer}.

In order to address the complexity of systems
 and unleash the potential of humans and technology,
 it is increasingly important to consider 
 technical aspects
 together with human, social, and organisational aspects~\citep{DBLP:journals/tasm/BucchiaroneDPCS20}.
In other words, a key challenge and opportunity
 revolves around the design of social machines~\citep{DBLP:conf/www/BuregioMR13,berners1999weaving-the-web}, 
hybrid societies~\citep{DBLP:journals/firai/HamannKBSFKMMRT16}, 
and socio-technical systems~\citep{DBLP:journals/iwc/BaxterS11}.
A social machine can be described as \emph{``a computational entity that blends computational and social processes''} and that is at the intersection of social software, people as computational units, and software as sociable entities~\citep{DBLP:conf/www/BuregioMR13,berners1999weaving-the-web}.
In this respect, elements whose formalisation and use 
 might promote the engineering of \ac{ci} into social machines
 may include macro-level and collective abstractions~\citep{DBLP:journals/tetc/ScekicSVRTMD20},
 social concepts~\citep{DBLP:journals/tasm/BellmanBHLMPST17},
 and coordination models~\citep{DBLP:journals/csur/MaloneC94}.
However, several challenges remain,
 related to 
 proper modelling of human computation,
 achieving effective communication and coordination between humans and machines,
 achieving self-improving system integration~\citep{DBLP:journals/fgcs/BellmanBDEGLLNP21}.


\revB{
\subsection{Summary of recommendations for future research on \ac{aci} engineering}\label{ssec:summary-recomms}

This section has discussed multiple issues and directions
  providing for plenty of research opportunities and challenges.  
To summarise, we recommend the following topics to be further investigated:
\begin{itemize}
\item language-based solutions to \ac{ci} programming, as also fostered by recent research on macro-programming~\citep{DBLP:journals/corr/abs-2201-03473,DBLP:journals/jisa/JuniorSBP21}, possibly also working as a foundation for explainability~\citep{DBLP:journals/corr/abs-2203-11547};

\item approaches and mechanisms for controlling or steering emergence and self-organisation~\citep{DBLP:journals/alife/GershensonTWS20,varenne2015programming-emergence}, together with efforts for building a deeper understanding of these very concepts (cf. \cite{gignoux2017emergence-graph-theory});

\item the role of \ac{ci} across the various level of modern computing systems (e.g., the application level, the middleware level, and the physical system level)~\citep{DBLP:journals/jisa/JuniorSBP21}, to address functional as well as non-functional aspects including, e.g., security, resilience, and resource efficiency;

\item designs for integrating manual and automatic approaches to \ac{ci} engineering, 
 for instance along the lines of \ac{marl} with specifications~\citep{DBLP:conf/icaart/RitzPMGSZWSSKL21a} or program synthesis~\citep{DBLP:conf/icml/BastaniIS20,aguzzi2022coord-ac-rl} of macro-programs;
 
 \item integration of human intelligence with machine intelligence into hybrid, collectively intelligent systems~\citep{smirnov2019human-machine-ci,DBLP:journals/ais/PeetersDBBNSR21}, e.g., leveraging wearable computing~\citep{DBLP:conf/huc/FerschaLZ14}, ways for combining methods for human teamwork with \ac{ai}, and self-organisation protocols considering both humans and artificial agents~\citep{smirnov2019human-machine-ci,DBLP:journals/tetc/ScekicSVRTMD20}.
\end{itemize}
Last but not least,
 we strongly believe that 
 the collective viewpoint
 has yet to find its place 
 within the software engineering practice.
Recent efforts on formal models and languages for \acp{cas}~\citep{DBLP:journals/sttt/NicolaJW20,DBLP:journals/jlap/ViroliBDACP19,DBLP:journals/tetc/ScekicSVRTMD20} might highlight a path in that direction.
}

\section{Conclusion}
\label{sec:conc}

\Acf{ci} is a rich theme 
 that builds on multi-, inter-, and trans-disciplinary collective endeavours.
However, research is largely fragmented across several \revB{specific research problems (cf. types of collective tasks), research methods (cf. manual vs. automatic \ac{ci} design), and even entire computer science research areas (cf. hybrid systems, \acp{cas}, \acp{mas}, etc.), 
 and comprehensive mapping studies are currently missing,}
 making it difficult 
 for people of \revB{diverse} backgrounds  
 to get a sense of the overall field
 and even a sense of \ac{ci}-related work
 in their sub-field.
This \revB{scoping review} aimed 
 at providing a comprehensive view on \ac{ci}
 for computer scientists and engineers,
 with emphasis on concepts and perspectives,
 and also providing some research highlights on the forms of \ac{ci} that most interest them, namely \acf{aci}, \acf{cci}, and \acf{hmci}.
The final part reviews some interesting opportunities and challenges for researchers in computer science and engineering.
These point at directions that,
 despite visionary and preliminary work, are yet to develop: \ac{ci} programming, integration of manual and automatic techniques for \ac{ci} engineering, integration of collectiveness and emergence, and hybrid human-machine systems.

\bibliographystyle{ACM-Reference-Format}
\bibliography{biblio}

\end{document}